\documentclass[letterpaper, 10 pt, conference]{ieeeconf}

\IEEEoverridecommandlockouts

\usepackage{cite}
\usepackage{amsmath,amssymb,amsfonts}
\usepackage{algorithmic}
\usepackage{graphicx}
\usepackage{textcomp}
\usepackage{xcolor}
\usepackage{siunitx}
\usepackage{float}
\usepackage[]{footmisc}
\usepackage{ar}
\usepackage{url}
\usepackage{balance}
\usepackage{nicefrac}

\newcommand{\bs}[1]{\boldsymbol{#1}}  
\newcommand{\ts}[1]{\text{#1}} 

\renewcommand{\baselinestretch}{0.829}

\def\BibTeX{{\rm B\kern-.05em{\sc i\kern-.025em b}\kern-.08em

    T\kern-.1667em\lower.7ex\hbox{E}\kern-.125emX}}

\begin{document}

\title{\LARGE \bf Feedback Control of a Single-Tail Bioinspired \mbox{$\bs{59}$-mg} Swimmer \\

\thanks{The research presented in this paper was partially funded by the Washington State University (WSU) Foundation and the Palouse Club through a Cougar Cage Award to \mbox{N.\,O.\,P.-A.}; the US National Science Foundation (NSF) through \mbox{Award\,\,2244082}; the Morgan Family Charitable Trust through a direct gift to \mbox{N.\,O.\,P.-A.}; and, the Koerner Family Foundation (KFF) through a graduate fellowship to \mbox{C.\,K.\,T.}. Additional funding was provided by the WSU Voiland College of Engineering and Architecture through a \mbox{start-up} fund to \mbox{N.\,O.\,P.-A.}.} %
\thanks{The authors are with the School of Mechanical and Materials Engineering, Washington State University (WSU), Pullman,\,WA\,99164,\,USA. Corresponding authors' \mbox{e-mail:} {\tt conor.trygstad@wsu.edu}~(C.\,K.\,T.); {\tt n.perezarancibia@wsu.edu} (N.\,O.\,P.-A.).}%
%}
}
\author{Conor\,K.\,Trygstad, Cody\,R.\,Longwell, Francisco\,M.\,F.\,R.\,Gon\c{c}alves, Elijah\,K.\,Blankenship, \\ and N\'estor\,O.\,P\'erez-Arancibia}

\maketitle
\thispagestyle{empty}
\pagestyle{empty}

\begin{abstract}
We present an evolved steerable version of the \mbox{single-tail} \textit{Fish-\&-Ribbon--Inspired Small Swimming Harmonic roBot} (\mbox{FRISSHBot}), a \mbox{$\bs{59}$-mg} biologically~inspired~swimmer, which is driven by a new \textit{\mbox{shape-memory alloy}}~\mbox{(SMA)-based} bimorph actuator. The new \mbox{FRISSHBot} is controllable in the \textit{\mbox{two-dimensional}}~\mbox{($\bs{2}$D)} space, which enabled the first demonstration of \mbox{feedback-controlled} trajectory tracking of a \mbox{single-tail} aquatic robot with onboard actuation at the subgram scale. These new capabilities are the result of a \mbox{physics-informed} design with an enlarged head and shortened tail relative to those of the original platform. Enhanced by its design, this new platform achieves forward swimming speeds of up to $\bs{13.6}\,\textbf{mm/s}$~($\bs{0.38}\,\textbf{Bl/s}$), which is over four times that of the original platform. Furthermore, when following \mbox{$\bs{2}$D} references in \mbox{closed~loop}, the tested \mbox{FRISSHBot} prototype attains forward swimming speeds of up to $\bs{9.1}\,\textbf{mm/s}$, \textit{root-mean-square} (RMS) tracking errors as low as \mbox{$\bs{2.6}$\,mm}, turning rates of up to \mbox{$\bs{13.1}$\,\textdegree{/s}}, and turning radii as small as \mbox{$\bs{10}$\,mm}.
\end{abstract}

\section{Introduction}
\label{SEC01}
\vspace{-0.5ex}
The recent development of numerous \mbox{mm-to-cm--scale} aquatic \mbox{robots\cite{TrygstadCK2024, BlankenshipEK2024, LongwellCR2024, ZhaoQ2021, WangZ2008, GuoS2003, DengX2005, KimB2005,   ChenY2015, ChenY2017, SpinoP2024,LiK2023, KimD2022, OzcanO2014, ZhangX2011, ZhaoD2025, TrygstadCK2023, SongYS2007, GaoH2024, ZhouS2017, ZeQ2022}}---with varying levels of performance and efficiency---indicates great interest and substantial progress toward the creation of \mbox{small-scale} autonomous swimmers and, foreseeably, swarms of microswimmers capable of assisting humans with complex tasks in \mbox{real-world} scenarios. Small swimmers have the potential to bring unique capabilities for aquatic applications where maneuvering in confined spaces is required; for example, the inspection of water distribution systems, in which pipe diameters can be as small as $50$ and \mbox{$100$\,mm} for residential and commercial bypass lines, \mbox{respectively\cite{TurnerJ2014}}. Additionally, many \mbox{cm-to-dm--scale} \mbox{fish-inspired} and \mbox{snake-inspired} swimmers have been recently \mbox{reported\cite{BerlingerF2021, BerlingerF2018, ShintakeJ2018, RossiC2011, BayatB2016, StruebigK2020}}. However, the scaling of those prototypes down to the \mbox{mm-scale} has yet to be accomplished. 

A gamut of propulsion mechanisms has been used to drive the \mbox{small-scale} aquatic robots mentioned here, including flapping tails and \mbox{fins\cite{ZhaoQ2021, WangZ2008, GuoS2003, DengX2005, KimB2005, BlankenshipEK2024, LongwellCR2024, TrygstadCK2024, ChenY2015, ChenY2017}}, \mbox{jets\cite{SpinoP2024,LiK2023}}, and pads and skates for surface crawling or \mbox{skating~\cite{KimD2022, OzcanO2014, ZhangX2011, ZhaoD2025, TrygstadCK2023, SongYS2007, GaoH2024, ZhouS2017}}. We envision the development of \mbox{insect-scale} \textit{autonomous underwater vehicles}~(AUVs) capable of operating robustly over a wide range of water depths. Flapping tails and jets are propulsion mechanisms that translate adequately to underwater environments, whereas crawling methods based on leveraging the surface tension of water do not. However, it has been reported that the efficiency of \mbox{jet-based} propulsion is two to three times lower than that of undulatory \mbox{propulsion\cite{WebberDM1986}}, which suggests that designs based on flapping tails and fins are the most advantageous choice for \mbox{insect-scale} AUV propulsion. 
\begin{figure}[t!]
\vspace{1.4ex}
\begin{center}
\includegraphics[width=0.48\textwidth]{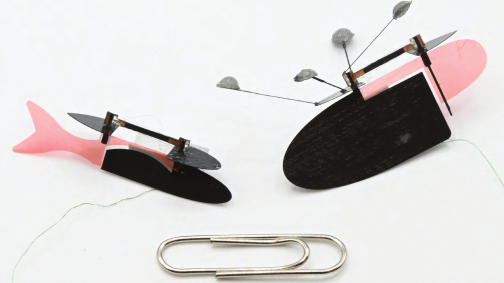}
\end{center}
\vspace{-2.0ex}    
\caption{\textbf{An evolved version of the \mbox{FRISSHBot} platform.}~A prototype of the original design presented \mbox{in\cite{TrygstadCK2024}} is shown on the left, and a prototype of the new design on the right. A \mbox{$28$-mm-long} paper clip was included in the photograph to indicate scale. \label{FIG01}}
\vspace{-2.0ex}
\end{figure}

Over the past few years, we have developed several \mbox{mm-scale} aquatic robots that employ undulating soft tails to generate the propulsive forces required for swimming by leveraging {\mbox{fluid--structure} interaction}~\mbox{\cite{BlankenshipEK2024,TrygstadCK2024,LongwellCR2024}}. The most notable of these prototypes is the \mbox{VLEIBot\textsuperscript{++}} presented \mbox{in\cite{LongwellCR2024}}, a \mbox{$900$-mg} autonomous swimmer that uses the \mbox{dual-tail} design introduced \mbox{in\cite{BlankenshipEK2024}} for generating thrust and achieving \textit{\mbox{two-dimensional}} ($2$D) tracking control. Nevertheless, considering the vast biological evidence available, we hypothesize that \mbox{single-tail} designs can potentially achieve higher swimming performance and efficiency; furthermore, \mbox{single-tail} robots can be designed to have relatively small \mbox{cross-sectional} areas, thus enabling high accessibility to confined spaces and maneuverability. These observations prompted us to develop the \mbox{$10$-mg} \textit{\mbox{shape-memory} alloy}~(SMA)-based bimorph actuator presented \mbox{in\cite{TrygstadCK2024}}, which is the key element that enabled the creation of the original \mbox{single-tail} \textit{Fish-\&-Ribbon--Inspired Small Swimming Harmonic roBot} (\mbox{FRISSHBot}), shown on the left in~\mbox{Fig.\,\ref{FIG01}}. 

Despite being the smallest robotic swimmer with onboard actuation developed at the time of publication, the swimming performance of the original \mbox{FRISSHBot} platform was suboptimal and steerability was not achieved. In this paper, we present an evolved controllable version of this swimmer---shown on the right in \mbox{Fig.\,\ref{FIG01}}. This new platform features modified head and tail designs, which were developed through basic \mbox{fluid-mechanics} analyses and collection of empirical data. The resulting swimmer has a mass of \mbox{$59$\,mg}, is \mbox{$36$\,mm} long, and achieves average swimming speeds on the order of $13.6\,\ts{mm/s}$~($0.38\,\ts{Bl/s}$), which is over four times faster than the original prototype. As discussed in \mbox{Section\,\ref{SEC04}}, this new platform is highly maneuverable and controllable in the \mbox{$2$D} space. Prior to this instance, only the subgram swimmer presented \mbox{in\cite{BlankenshipEK2024}} has been demonstrated to operate in closed loop, employing onboard actuation. 

Here, we present \mbox{closed-loop} \mbox{trajectory-tracking} experiments in which the tested \mbox{FRISSHBot} prototype achieved control errors with \textit{\mbox{root-mean-square}}~(RMS) values as low as \mbox{$2.6$\,mm} during rectilinear locomotion---an improvement of \mbox{$35$\hspace{0.1ex}$\%$} with respect to that achieved by the \mbox{dual-tail} VLEIBot\textsuperscript{+} platform presented \mbox{in\cite{BlankenshipEK2024}}. During \mbox{closed-loop} \mbox{left-turn} and \mbox{right-turn} tracking experiments, the tested swimmer achieved turning rates of up to $13.1$\,\textdegree{/s}, which is comparable to the rates reported for the \mbox{VLEIBot\textsuperscript{+}} \mbox{in\cite{BlankenshipEK2024}}. Notably, the measured turning radii during these tests are as small as \mbox{$10$\,mm}~(\mbox{$0.28$\,Bl}), which is about \mbox{$88$\hspace{0.1ex}$\%$} lower than that measured for the \mbox{VLEIBot\textsuperscript{+}}. These new results provide empirical evidence suggesting that a higher swimming maneuverability can be attained using a \mbox{single-tail} propulsor in comparison to that achievable with \mbox{dual-tail} designs. To the best of our knowledge, this is the first demonstration of \mbox{feedback-controlled} swimming tests employing a \mbox{single-tail} subgram platform reported to date.

The rest of the paper is organized as follows. \mbox{Section\,\ref{SEC02}} presents the \mbox{physics-based} design and functionality of the new \mbox{FRISSHBot} platform. \mbox{Section\,\ref{SEC03}} describes the experiments performed to characterize the \mbox{peak-to-peak} lateral excursion of the swimmer's tail during actuation, and presents an analysis of the measured forward speeds and turning rates across its range of operation. \mbox{Section\,\ref{SEC04}} presents \mbox{feedback-controlled} \mbox{trajectory-tracking} swimming tests using the new \mbox{FRISSHBot} platform and discusses the achieved performance in terms of tracking errors. Last, \mbox{Section\,\ref{SEC05}} summarizes the research presented in the paper and discusses directions for future work.
\begin{figure}[t!]
\vspace{1.4ex}    
\begin{center}    
\includegraphics[width=0.48\textwidth]{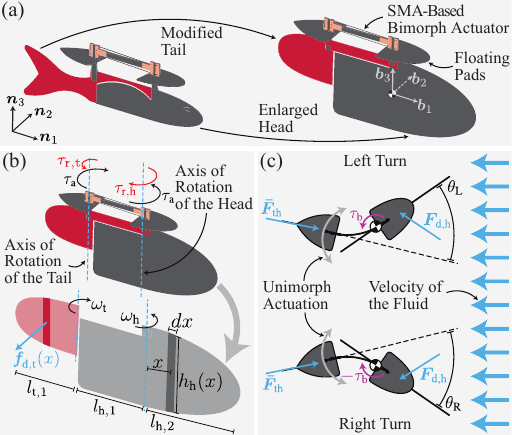}
\end{center}
\vspace{-2ex}
\caption{\textbf{Design and functionality of the new \mbox{FRISSHBot} platform.} The vector triplets \mbox{$\left\{\bs{n}_1,\bs{n}_2,\bs{n}_3 \right\}$} and \mbox{$\left\{\bs{b}_1,\bs{b}_2,\bs{b}_3 \right\}$} define the inertial and \mbox{body-fixed} frames used for kinematic and kinetic analyses.\,\textbf{(a)}\,Design evolution of the \mbox{FRISSHBot} platform. As seen, the planform of the tail was modified from a caudal to a parabolic shape, and the planform of the head was enlarged.\,\textbf{(b)}\,Idealized dynamical model of the new \mbox{FRISSHBot} platform. During swimming, the robot's actuator generates two torques with equal instantaneous signed magnitudes, $\tau_{\ts{a}}(t)$, and opposite directions along the $\bs{b}_3$ axis. The periodic oscillation of the swimmer generates resistive hydrodynamic torques acting on the head and tail; the signed magnitudes of these torques are labeled as $\tau_{\ts{r,h}}$ and $\tau_{\ts{r,t}}$, respectively.\,\textbf{(c)}\,Turning during swimming is achieved by operating the robot's actuator in a unimorph mode---left~or~right. A left turn is generated by the aggregation of two torques with direction $\bs{b}_3$---one \mbox{tail-induced} and the other \mbox{head-induced}. Similarly, a right turn is generated by the aggregation of two torques with direction $-\bs{b}_3$---one \mbox{tail-induced} and the other \mbox{head-induced}. \label{FIG02}}
\vspace{-2.2ex}
\end{figure}

\section{Design of the New FRISSHBot}
\label{SEC02}
\vspace{-0.5ex}
In this section, we present the design and discuss the functionality of the new FRISSHBot platform. As seen in \mbox{Fig.\,\ref{FIG02}(a)}, we evolved the design of the original swimmer presented \mbox{in~\cite{TrygstadCK2024}} to increase locomotion performance and achieve empirical \mbox{$2$D} controllability. The two main modifications are an enlarged head and a new propulsive tail, which we varied from a caudal to a parabolic planform. It is a \mbox{well-established} empirical fact that, for an underwater propulsive flapping foil operating at low Reynolds numbers, greater stroke amplitudes and frequencies typically generate higher thrust \mbox{forces~\cite{DingH2024,JamesCR2022}}. Prompted by this physical fact, we aimed to generate a design that increases the stroke amplitude of the swimmer's tail compared to that of the original robot---which we use as a benchmark---at the same frequency of actuation. For analysis and development purposes, we use the dynamical model in~\mbox{Fig.\,\ref{FIG02}(b)}. As seen here, a \mbox{FRISSHBot} prototype is mechanically powered by a bimorph \mbox{SMA-based} actuator of the type introduced \mbox{in~\cite{TrygstadCK2024}}. 

To describe the kinematics and kinetics of the system, we use the \mbox{laboratory-fixed} inertial and \mbox{body-fixed} frames of reference defined by the vector triplets \mbox{$\left\{\bs{n}_1,\bs{n}_2,\bs{n}_3 \right\}$} and \mbox{$\left\{\bs{b}_1,\bs{b}_2,\bs{b}_3 \right\}$}, respectively; the origin of the \mbox{body-fixed} frame coincides with the \textit{center of mass} (CoM) of the robot. According to the model in \mbox{Fig.\,\ref{FIG02}(b)}, during swimming, the actuator generates two torques with equal instantaneous signed magnitudes, $\tau_{\ts{a}}(t)$, and opposite directions along the $\bs{b}_3$ axis. Note that, as defined, \mbox{$\tau_{\ts{a}}(t) > 0$} when the corresponding vector, \mbox{$\bs{\tau}_{\hspace{-0.2ex}\ts{a}}(t)$}, acting on the head points in the direction of $\bs{b}_3$. Consistent with this convention and the graphically defined vectors in~\mbox{Fig.\,\ref{FIG02}(b)}, the instantaneous signed angular speed of the head, $\omega_{\ts{h}}(t)$, is positive when the direction of its corresponding angular velocity, $\bs{\omega}_{\ts{h}}(t)$, points in the direction of $\bs{b}_3$; similarly, the instantaneous signed angular speed of the tail, $\omega_{\ts{t}}(t)$, is positive when the direction of its corresponding angular velocity, $\bs{\omega}_{\ts{t}}(t)$, points in the direction of $-\bs{b}_3$. As described \mbox{in\cite{TrygstadCK2024}}, the bimorph actuator driving the swimmer can be thought of as being composed of two \mbox{SMA-based} subsystems---left~and~right. When both sides are actuated alternatingly, we say that the actuator functions in the bimorph mode; when only one side is actuated, we say that the actuator functions in a unimorph mode---left~or~right. In the bimorph mode, the swimmer oscillates by cyclically bending to the left and right, while, by design, its head functions as an anchor and its tail as a propulsor.
\begin{figure*}[t!]
\vspace{1.4ex}
\begin{center}
\includegraphics[width=0.96\textwidth]{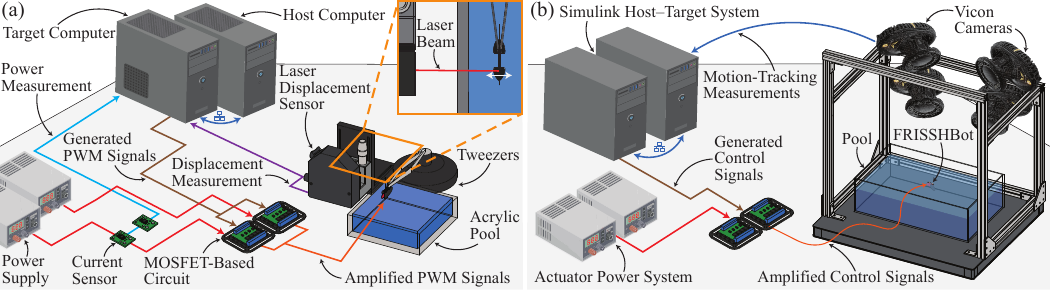}
\end{center}
\vspace{-2ex}    
\caption{\textbf{Experimental characterization of the new \mbox{FRISSHBot} platform.}\,\textbf{(a)}\,Setup used to measure the lateral excursion of the swimmer's tail and power consumption of the driving actuator during constrained operation. In this setup, signals are generated, monitored, and recorded using a Mathworks Simulink \mbox{Real-Time} \mbox{host--target} system equipped with a \mbox{PCI-$6229$} \mbox{AD/DA} board. The two PWM voltages used to excite the bimorph actuator of the tested prototype are generated following the method described \mbox{in~\cite{TrygstadCK2024}}.\,\textbf{(b)}\,Setup used to perform swimming experiments with the tested \mbox{FRISSHBot} prototype. The \mbox{host--target} system in~(a) is used for signal generation, monitoring, data collection, and control. A Vicon motion-capture system with four \mbox{VK$16$} cameras measures and records the position and orientation of the robot in space, which is required for \mbox{closed-loop} control implementation. \label{FIG03}}
\vspace{-2.0ex}
\end{figure*}

By mathematically describing the head and tail of the robot as plates, we can model the aggregated viscous and inertial reactions of the surrounding water acting on the swimmer as two reactive torques: one corresponding to the head, with signed magnitude $\tau_{\ts{r},\ts{h}}$, which is positive when the direction of its corresponding vector, $\bs{\tau}_{\hspace{-0.2ex}\ts{r},\ts{h}}$, points in the direction of $-\bs{b}_3$; and, another corresponding to the tail, with signed magnitude $\tau_{\ts{r},\ts{t}}$, which is positive when the direction of its corresponding vector, $\bs{\tau}_{\hspace{-0.2ex}\ts{r},\ts{t}}$, points in the direction of $\bs{b}_3$. Thus, noting that in this case all the torques are applied along the direction of $\bs{b}_3$ or $-\bs{b}_3$, the instantaneous total torque acting on the robot during swimming, $\bs{\tau}_{\hspace{-0.2ex}\ts{b}}(t) = \tau_{\ts{b}}(t) \cdot \bs{b}_3$, can be described with the scalar equation

\vspace{-2ex}
{\small
\begin{align}
\begin{split}
\tau_{\ts{b}}(t) &= \tau_{\ts{h}}(t) + \tau_{\ts{t}}(t) \\
&= \left[\tau_{\ts{a}}(t) + \tau_{\ts{r},\ts{h}}(t) \right] +
\left[-\tau_{\ts{a}}(t) + \tau_{\ts{r},\ts{t}}(t) \right] \\
&= - \tau_{\ts{r},\ts{h}}(t) + \tau_{\ts{r},\ts{t}}(t),
\end{split}
\label{EQN01}
\end{align}
}

\noindent where $\tau_{\ts{h}}(t)$ and $\tau_{\ts{t}}(t)$ are the instantaneous signed magnitudes of the total torques acting on the head and tail, respectively. Since the robot is assumed to swim rectilinearly, its net time-average rotation, with respect to the inertial frame, after an actuation cycle must be zero. Thus, from Euler's second law,

\vspace{-2ex}
{\small
\begin{align}
\langle \tau_{\ts{b}}(t) \rangle = \langle - \tau_{\ts{r},\ts{h}}(t) + \tau_{\ts{r},\ts{t}}(t) \rangle =
-\langle \tau_{\ts{r},\ts{h}}(t) \rangle + \langle \tau_{\ts{r},\ts{t}}(t) \rangle = 0,
\label{EQN02}
\end{align}
}

\noindent from which it follows that

\vspace{-2ex}
{\small
\begin{align}
\langle \tau_{\ts{r},\ts{h}}(t) \rangle = \langle \tau_{\ts{r},\ts{t}}(t) \rangle.
\label{EQN03}
\end{align}
}

For each of the two plates, using the method \mbox{in~\cite{NakayamaY2000}}, we can compute the magnitude of the corresponding total reactive torque as 

\vspace{-2ex}
{\small
\begin{align}
\begin{split}
\tau_{\ts{r},\ts{p}}(t) &= \int_{-l_{\ts{p},1}}^{l_{\ts{p},2}} f_{\ts{d},\ts{p}}(x)x dx \\
&= -\frac{1}{2} \rho C_{\ts{d}} \omega_{\ts{p}}(t) \left| \omega_{\ts{p}}(t) \right| \int_{-l_{\ts{p},1}}^{l_{\ts{p},2}} h_{\ts{p}}(x) \left| x \right|^3dx,
\end{split}
\label{EQN04}
\end{align}
}

\noindent in which \mbox{$\ts{p} \in \left\{ \ts{h}, \ts{t} \right\}$}; \mbox{$-l_{\ts{p},1} \leq 0$} and \mbox{$l_{\ts{p},2} \geq 0$} are the lateral limits of the plate, measured from the axis of rotation; \mbox{$f_{\ts{d},\ts{p}}(x) =-  \frac{1}{2} \rho C_{\ts{d}} h_{\ts{p}}(x) \omega_{\ts{p}}(t) \left| \omega_{\ts{p}}(t) \right| x \left|x\right|$} is the signed magnitude of the drag force per unit length along the span of the plate at a distance $x$ from the axis of rotation; $\rho$ is the density of water; $C_{\ts{d}}$ is a dimensionless drag coefficient; $\omega_{\ts{p}}(t)$ is the instantaneous signed angular speed at which the plate rotates about its axis of rotation; and, $h_{\ts{p}}(x)$ is the plate chord at position $x$. Thus, substituting (\ref{EQN04}) into (\ref{EQN03}), we obtain that, during rectilinear swimming,

\vspace{-2ex}
{\small
\begin{align}
\langle \omega_{\ts{h}}^2(t) \rangle  \int_{-l_{\ts{h},1}}^{l_{\ts{h},2}} h_{\ts{h}}(x) \left| x \right|^3dx  =
\langle \omega_{\ts{t}}^2(t) \rangle  \int_{-l_{\ts{t},1}}^{l_{\ts{t},2}} h_{\ts{t}}(x) \left| x \right|^3dx,
\label{EQN05}
\end{align}
}

\noindent in which \mbox{$I_{\ts{h}} =  \int_{-l_{\ts{h},1}}^{l_{\ts{h},2}} h_{\ts{h}}(x) \left| x \right|^3dx$}, \mbox{$I_{\ts{t}} = \int_{-l_{\ts{t},1}}^{l_{\ts{t},2}} h_{\ts{t}}(x) \left| x \right|^3dx$} are \textit{resistive drag factors} (RDFs). Note that in the derivation of the expression specified by (\ref{EQN05}), we used the fact that, in agreement with the graphical definitions in~\mbox{Fig.\,\ref{FIG02}(b)} and the dynamics of the \mbox{FRISSHBot}, at any given instant, the signs of $\omega_{\ts{h}}(t)$ and $\omega_{\ts{t}}(t)$ are the same. 
\begin{figure*}[t!]
\vspace{1.4ex}
\begin{center}  
\includegraphics[width=0.96\textwidth]{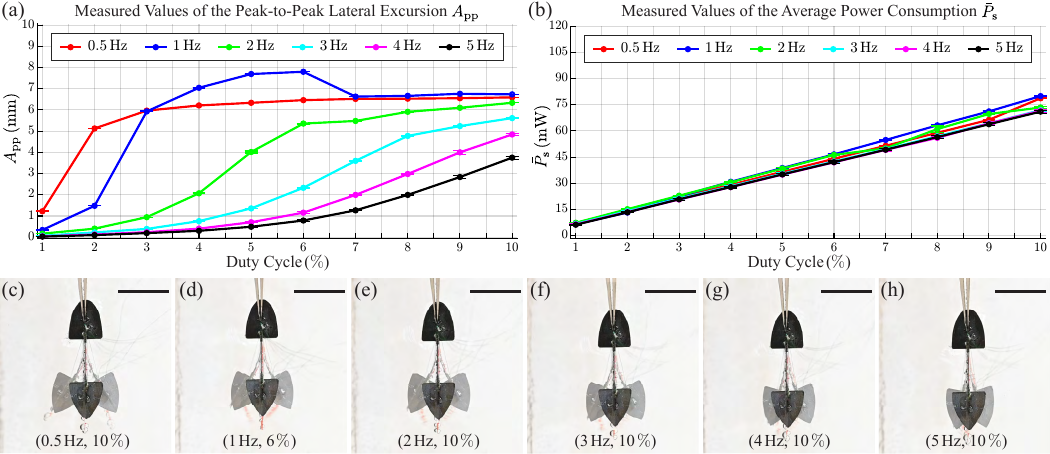}
\end{center}
\vspace{-2.0ex}
\caption{\textbf{\mbox{Peak-to-peak} lateral excursion and average power consumption.}\,\textbf{(a)}\,Each data point corresponds to the mean---and associated ESD---of the measured \mbox{peak-to-peak} lateral excursions, $A_{\ts{pp}}$, in \mbox{$30$\,s} of \mbox{steady-state} data. In this case, each ESD value represents the \mbox{cycle-to-cycle} variation of the excursion measurements for the corresponding experiment.\,\textbf{(b)}\,Each data point corresponds to the mean---and associated ESD---of the average power consumption, $\bar{P}_{\ts{s}}$, from \mbox{$30$\,s} of \mbox{steady-state} data. In this case, each ESD value represents the \mbox{cycle-to-cycle} variation of the average power consumption.\,\mbox{\textbf{(c)--(h)}}\,Composites of video frames taken at the instants of maximum stroke displacements (left and right), and at rest for PWM excitations with frequencies of $0.5$, $1$, $2$, $3$, $4$, and \mbox{$5$\,Hz}, while operating at their respective DC values of maximum lateral excursion. The scale bars in the images indicate a length of \mbox{$10$\,mm}. Video footage of these experiments is shown in the accompanying supplementary movie. \label{FIG04}}
\vspace{-1.8ex}
\end{figure*}

The expression specified by (\ref{EQN05}) can be used directly as a guideline for design: specifically, to maximize the angular speed of the tail during swimming---such that it functions as a propulsor---and minimize the angular speed of the head---such that it functions as an anchor. This expression indicates that a simple approach to increase the value of $I_{\ts{h}}$, relative to that of $I_{\ts{t}}$, is to make the head taller and longer than the tail. In the design of the original \mbox{FRISSHBot} platform, the head and tail have similar average heights, i.e., \mbox{$\bar{h}_{\ts{h}} \approx \bar{h}_{\ts{t}}$}. Additionally, the tail of that platform was designed with most of its area located far away from its axis of rotation, which resulted in a relatively high $I_{\ts{t}}$ value and corresponding reactive torque, thus limiting the tail's mobility relative to the head and surrounding water. We hypothesize that this lack of mobility severely degrades the swimming performance of that robot. As shown in \mbox{Figs.\,\ref{FIG02}(a)}~and~(b), in the design of the new \mbox{FRISSHBot}, the head is significantly taller along the vertical dimension than the tail, which is shorter along the longitudinal dimension than in the original robot, thereby placing the tail's center of area closer to its axis of rotation. Together, these two design modifications resulted in a significantly larger reactive torque acting on the head than on the tail. We determined the final dimensions of the robotic design---particularly those of the head and tail---using a heuristic process based on \mbox{open-loop} \mbox{straight-swimming} and turning experiments, with the goal of maximizing speed and maneuverability. The resulting values for the RDFs of the new design are \mbox{$I_{\ts{h}} = 1.14\times10^5\,\ts{mm}^5$} and \mbox{$I_{\ts{t}} = 1.07\times10^4\,\ts{mm}^5$}. The values for the RDFs of the old design are \mbox{$I_{\ts{h}} = 1.88\times10^4\,\ts{mm}^5$} and \mbox{$I_{\ts{t}} = 2.19\times10^4\,\ts{mm}^5$}. Thus, through the design process, we modified the robot such that \mbox{$I_{\ts{h}}: 1.88 \times 10^4\,\ts{mm}^5 \rightarrow 1.14 \times 10^5\,\ts{mm}^5$} and \mbox{$I_{\ts{t}}: 2.19 \times 10^4\,\ts{mm}^5 \rightarrow 1.07 \times 10^4\,\ts{mm}^5$}.

Once more employing (\ref{EQN05}) for basic dynamical analysis, it can be determined that the enhanced design shown in~\mbox{Figs.\,\ref{FIG02}(a)}~and~(b) not only increases the swimming performance in terms of \mbox{steady-state} speed, but also improves the performance and reliability of turning, thereby enabling full \mbox{$2$D} controllability. An explanation of the turning mechanism and steering strategy is depicted in \mbox{Fig.\,\ref{FIG02}(c)}. As seen here, to perform a turn, the actuator of the swimmer is operated in one of the two unimorph modes, as described \mbox{in~\cite{TrygstadCK2024}}. When operated in a unimorph mode, the robot periodically bends asymmetrically with a bias in the direction of actuation. Namely, when the left \mbox{SMA-based} subsystem of the actuator is excited and the right subsystem is kept inactive, the robot arcs periodically with its distal ends toward the left and its point of maximum convexity toward the right. Thus, during each actuation cycle, the lateral excursion of the tail relative to the longitudinal axis of the head is biased toward the left, which is denoted as $\theta_{\ts{L}}$ in \mbox{Fig.\,\ref{FIG02}(c)}. Analogously, when the right \mbox{SMA-based} subsystem of the actuator is excited and the left subsystem is kept inactive, the robot arcs periodically with its distal ends toward the right and its point of maximum convexity toward the left. As a consequence, during each actuation cycle, the lateral excursion of the tail relative to the longitudinal axis of the head is biased toward the right, which is denoted as $\theta_{\ts{R}}$ in \mbox{Fig.\,\ref{FIG02}(c)}.

During a left turn, the biased periodic bending of the swimmer forces the average thrust vector produced by the tail, $\bs{\bar{F}}_{\hspace{-0.3ex}\ts{th}}$, to act at an angle $\theta_{\ts{L}}$ (see the upper illustration in \mbox{Fig.\,\ref{FIG02}(c)}) with respect to the longitudinal axis of the head. By design, this vector also acts at a distance from the robot's CoM, thus creating a \mbox{tail-induced} \mbox{left-turning} torque (about $\bs{b}_3$) and, as a result, a left turn. Additionally, since, immediately prior to initiating a turn, the robot is swimming forward in the direction of $\bs{b}_1$, the \mbox{tail-induced} left turn generates an additional hydrodynamic drag force, $\bs{F}_{\hspace{-0.3ex}\ts{d},\ts{h}}$---with inertial and viscous components---that, in turn, produces a \mbox{head-induced} \mbox{left-turning} torque (about $\bs{b}_3$), which increases the rate of left rotation. During a right turn, the biased periodic bending of the swimmer forces the average thrust produced by the tail, $\bs{\bar{F}}_{\hspace{-0.3ex}\ts{th}}$, to act at an angle $\theta_{\ts{R}}$ (see the lower illustration in \mbox{Fig.\,\ref{FIG02}(c)}) with respect to the longitudinal axis of the head. By design, this thrust also acts at a distance from the robot's CoM, thus creating a \mbox{tail-induced} \mbox{right-turning} torque (about $-\bs{b}_3$) and, as a result, a right turn. Furthermore, analogous to the \mbox{left-turning} case, a \mbox{tail-induced} right turn induces an additional hydrodynamic drag force, $\bs{F}_{\hspace{-0.3ex}\ts{d},\ts{h}}$---with inertial and viscous components---that, in turn, generates a \mbox{head-induced} \mbox{right-turning} torque (about $-\bs{b}_3$), which increases the rate of right rotation.

In summary, a left turn is generated by the aggregation of two torques with direction $\bs{b}_3$---one \mbox{tail-induced} and the other \mbox{head-induced}. Similarly, a right turn is generated by the aggregation of two torques with direction $-\bs{b}_3$---one \mbox{tail-induced} and the other \mbox{head-induced}. By leveraging the enhanced capabilities of the new \mbox{FRISSHBot} design, we demonstrate that \mbox{$2$D} controllability can be achieved through a combination of unimorph and bimorph actuation modes. 

\section{Characterization of the New FRISSHBot}
\label{SEC03}
\vspace{-0.5ex}
\subsection{Metrics of Swimming Efficiency}
\label{SUBSEC03A}
\vspace{-0.5ex}
Two common metrics employed to compare the efficiency of swimming robots are the \textit{\mbox{cost~of~transport}}~(CoT) and \textit{Strouhal~number}~(St), defined as

\vspace{-2ex}
{\small
\begin{align}
\ts{CoT} = \frac{\bar{P}_{\ts{s}}}{mg \bar{v}} \quad \ts{and} \quad \ts{St} = \frac{f_{\ts{o}}A_{\ts{pp}}}{\bar{v}}, 
\label{EQN06}
\end{align}
}

\noindent in which $\bar{P}_{\ts{s}}$ is the average power consumed by the swimmer during operation; $m$ is the total mass of the swimmer; $g$ is the acceleration of gravity; $\bar{v}$ is the average speed of the swimmer in steady state, relative to the surrounding water; $f_{\ts{o}}$ is the frequency at which the swimmer's tail is oscillated during operation; and, $A_{\ts{pp}}$ is the \mbox{peak-to-peak} lateral excursion---left to right or vice versa, as shown in \mbox{Fig.\,\ref{FIG02}(b)}---of the tail at its free distal end. Recently, however, another metric of swimming efficiency---the \textit{swim number}~(Sw)\cite{CoeM2024}---has been proposed because it provides a unified scaling law across all swimming modes and size scales. Namely,

\vspace{-2ex}
{\small
\begin{align}
\ts{Sw} = 2\pi \cdot \ts{Re} \cdot \ts{St} = 2 \pi \cdot \frac{\bar{v} L}{\nu} \cdot \frac{f_{\ts{o}} A_{\ts{pp}}}{\bar{v}} = \frac{2\pi f_{\ts{o}}A_{\ts{pp}}L}{\nu},
\label{EQN07}
\end{align}
}

\noindent in which $\ts{Re}$ is the Reynolds number; $L$ is the length of the swimmer; and, $\nu$ is the kinematic viscosity of water.

Directly from its definition, it follows that the CoT quantifies the energy required to move a unit of weight over a unit of distance, which implies that this parameter obeys different scaling laws across swim modes and size scales. Along the same lines, directly from its definition, it follows that the St---by relating tail-beat frequency, amplitude, and forward speed---provides insight into propulsion efficiency and flow regime. A limitation of this definition is that the St value quantifies swimming performance using only one length scale, which limits its ability to reflect the complex, multi-scale nature of real swimming systems---especially when comparing across swim modes or size scales. In contrast, the Sw accounts for both relevant length scales---oscillation amplitude and body length---as well as the fluid properties. To experimentally quantify the CoT, St, and Sw, we measured the values of $\bar{P}_{\ts{s}}$, $\bar{v}$, and $A_{\ts{pp}}$ across a range of \textit{\mbox{pulse-width} modulation} (PWM) excitation signals, using the experimental setups discussed next.
\begin{figure}[t!]
\vspace{1.4ex}
\begin{center}    
\includegraphics[width=0.48\textwidth]{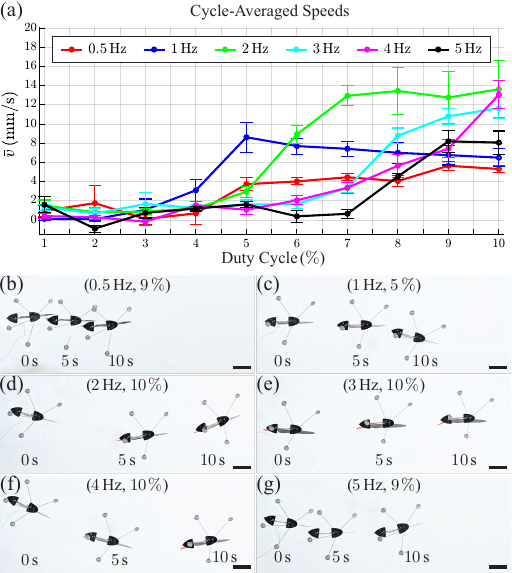}
\end{center}    
\vspace{-2ex}
\caption{\textbf{Average speed values recorded during \mbox{open-loop} swimming tests.}\,\textbf{(a)}\,Each data point corresponds to the mean---and associated ESD---of the \mbox{cycle-averaged} swimming speeds measured from \mbox{$10$\,s} of \mbox{steady-state} data. In this case, each ESD value represents the \mbox{cycle-to-cycle} variation of the average speed measurements.\,\textbf{(b)--(g)}\,Composites of frames taken at \mbox{$5$-s} intervals from overhead video footage of the \mbox{FRISSHBot} prototype swimming at its maximum recorded speed for each tested frequency reported in~(a). The scale bars in the images indicate a length of \mbox{$10$\,mm}. Video footage of these experiments is shown in the accompanying supplementary movie. \label{FIG05}}
\vspace{-1.76ex}
\end{figure}

\subsection{Experimental Setups}
\label{SUBSEC03B}
\vspace{-0.5ex}
We use the experimental setup depicted in~\mbox{Fig.\,\ref{FIG03}(a)} to measure the lateral excursion of the tail and power consumed by the tested \mbox{FRISSHBot} prototype during constrained operation (see the inset on the \mbox{upper-right}). In this setup, signals are generated, monitored, and recorded using a Mathworks Simulink \mbox{Real-Time} \mbox{host--target} system equipped with a \mbox{PCI-$6229$} \mbox{AD/DA} board. The two PWM voltages used to excite the bimorph actuator of the tested swimmer are generated employing the method described \mbox{in~\cite{TrygstadCK2024}}. Accordingly, both exciting signals have the same frequency, $f$---which in this case coincides with $f_{\ts{o}}$---and the same \mbox{on-height}, but their phases are shifted by \mbox{$180$\textdegree}. To supply the power consumed by the two \mbox{SMA-based} subsystems of the actuator, we use two \mbox{MOSFET-based} switching circuits (\mbox{four-channel} \mbox{YYNMOS-$4$}). During symmetric bimorph actuation, the \textit{duty~cycle}~(DC) of both exciting PWM signals is identical; during unimorph actuation, as explained in Section\,\ref{SEC02}, one side of the actuator is excited while the other side is left inactive \mbox{($\ts{DC}=0$)}. A diverse set of mixed actuation modes---created by sequentially combining bimorph and unimorph modes---can be generated by varying the DC values of the PWM signals exciting both sides of the driving actuator. As seen on the \mbox{bottom-left} of the illustration in \mbox{Fig.\,\ref{FIG03}(a)}, two current sensors (\mbox{Pololu\,$05$AU})---connected in series with the power supplies that feed the \mbox{MOSFET-based} circuits---are used to measure the instantaneous currents flowing through each side of the actuator during operation. Similarly to the cases discussed \mbox{in\cite{TrygstadCK2023,NguyenXT2020}},~and~\cite{BenaRM2021}, during the tests, we measure the instantaneous displacement of the actuator at the connection point with the robot's tail using a laser displacement sensor (\mbox{Keyence\,LK-$031$}). All the signals used for system characterization are generated and recorded at a \mbox{sample-and-hold} rate of \mbox{$10$\,kHz}.

Before conducting the experimental characterization of the lateral excursion of the robot's tail and power consumption of the actuator during operation, we connected five \mbox{$52$-AWG} tether wires in parallel to each terminal on both sides of the bimorph actuator. This electrical connection was heuristically selected to minimize power dissipation while maintaining sufficient flexibility to avoid any impediment to the functionality of the actuator. In addition, we prepared a swimming environment for the robot by filling a small acrylic pool with distilled water. During these experiments, the head of the tested \mbox{FRISSHBot} prototype is held immobile using a pair of tweezers, in the same position relative to the water surface as during swimming---i.e., with the floating pads in contact with the surface---as shown in the inset of \mbox{Fig.\,\ref{FIG03}(a)}. For a consistent basis of comparison and analysis, we heuristically set the \mbox{on-height} voltage of the exciting PWM signals such that a nominal current of \mbox{$250$\,mA} flows through each side of the actuator; in this case, the required \mbox{on-height} voltage is approximately \mbox{$4$\,V}. For the lateral excursion characterization, we excited the bimorph actuator of the tested prototype using all the PWM signals corresponding to the \mbox{$f$--DC} pairs with values in the sets \mbox{$f\in\{0.5,1,2,3,4,5\}$}\,Hz and \mbox{$\ts{DC}\in\{1\hspace{-0.4ex}:\hspace{-0.4ex}1\hspace{-0.4ex}:\hspace{-0.4ex}10\}$\hspace{0.1ex}$\%$}. For each data set corresponding to a tested PWM excitation, we computed the mean of the measured \mbox{peak-to-peak} tail lateral excursions and mean of the robot's \mbox{cyclic-averaged} power consumptions from \mbox{$30$\,s} of \mbox{steady-state} data. Additionally, for each mean value, we computed the corresponding \textit{experimental standard deviation}~(ESD), which quantifies the \mbox{cycle-to-cycle} variation in lateral excursion of the tail and average power consumption of the robot. Note that by immobilizing the head of the swimmer during the tests, we maximize the \mbox{peak-to-peak} lateral excursion of the tail; thus, the estimates of the St and Sw---presented in Section\,\ref{SUBSEC03C}---should be considered to be upper bounds.

To assess the swimming speed and turning performance of the tested \mbox{FRISSHBot} prototype, we used the experimental setup depicted in \mbox{Fig.\,\ref{FIG03}(b)}, in which the same \mbox{host--target} and \mbox{power-electronics} systems in \mbox{Fig.\,\ref{FIG03}(a)} are used to drive the bimorph actuator of the swimmer. During these swimming tests, a Vicon motion-capture system with four \mbox{VK$16$} cameras measures and records the position and orientation of the robot in space. These data are streamed, employing \mbox{Tracker\,$10.0$}, to the \mbox{host--target} system at a rate of \mbox{$250$\,Hz} for monitoring, feedback control, and collection. Before performing a swimming experiment, we install retroreflective markers on the tested swimmer for motion tracking and place the robot in a pool filled with distilled water. For assessing \mbox{open-loop} forward swimming, we used the same set of exciting PWM signals as in the \mbox{lateral-excursion} experiments discussed above. During \mbox{open-loop} turning experiments---performed using the strategy described in \mbox{Section\,\ref{SEC02}} and shown in \mbox{Fig.\,\ref{FIG02}(c)}---we observed that PWM excitations with frequencies below \mbox{$1$\,Hz} or DC values below \mbox{$5$\hspace{0.1ex}$\%$} do not produce noticeable turning. Therefore, we only tested unimorph actuation for turning---in both directions---using PWM excitations defined by the \mbox{$f$--DC} pairs with values in the sets \mbox{$f\in\{1,2,3,4,5\}$}\,Hz and \mbox{$\ts{DC}\in\{5\hspace{-0.2ex}:\hspace{-0.2ex}1\hspace{-0.2ex}:\hspace{-0.2ex}15\}$\hspace{0.1ex}$\%$}. For each tested combination of PWM voltages, we estimated the average signed speed of the tested \mbox{FRISSHBot} prototype---along the $\bs{b}_{1}$ axis---and average signed turning rate---about the $\bs{b}_3$ axis---from \mbox{$10$\,s} of \mbox{steady-state} experimental data. The results for these experiments are discussed next.
\begin{figure*}[t!]
\vspace{1.4ex}
\begin{center}
\includegraphics[width=0.96\textwidth]{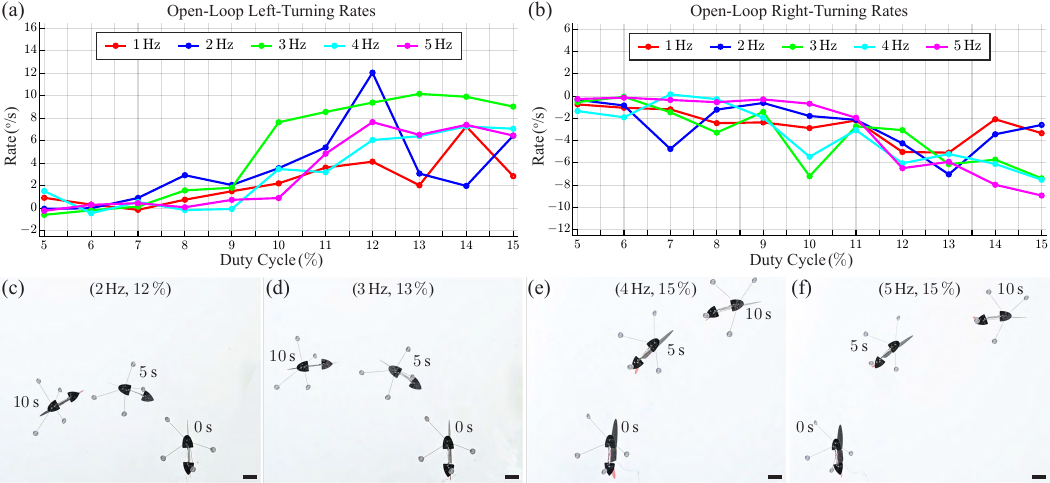}
\end{center}
\vspace{-2.0ex}    
\caption{\textbf{\mbox{Open-loop} turning maneuvers.}\,\textbf{(a)--(b)}\,Each data point indicates the average signed turning rate (relative to $\bs{b}_3$) recorded during a turning maneuver, computed from \mbox{$10$\,s} of \mbox{steady-state} data.\,\textbf{(c)--(f)}\,Composites of frames taken at \mbox{$5$-s} intervals from overhead video footage of the \mbox{FRISSHBot} prototype performing turns at its maximum recorded turning rate as reported in (a)~and~(b). The scale bars in the images indicate a length of \mbox{$10$\,mm}. Video footage of these experiments is shown in the accompanying supplementary movie. \label{FIG06}}
\vspace{-1.70ex}
\end{figure*}

\subsection{Results and Discussion}
\label{SUBSEC03C}
\vspace{-0.5ex}
\mbox{Fig.\,\ref{FIG04}(a)} summarizes the experimental data obtained from the tests conducted to measure the \mbox{peak-to-peak} lateral excursion, $A_{\ts{pp}}$, of the flapping tail for the selected set of PWM excitations (see Section\,\ref{SUBSEC03B}). \mbox{Fig.\,\ref{FIG04}(b)} summarizes the data on the average power consumption corresponding to the same experiments. In the plot of \mbox{Fig.\,\ref{FIG04}(a)}, each data point shows the mean value and corresponding ESD of the measured cyclic $A_{\ts{pp}}$ values in \mbox{$30$\,s} of \mbox{steady-state} data. As observed, the \mbox{peak-to-peak} lateral excursion increases with the DC value. At an actuation frequency of $1$\,Hz, the average of the lateral excursion reaches its maximum of \mbox{$7.80$\,mm} at a DC of \mbox{$6$\hspace{0.1ex}$\%$}, then slightly decreases at higher DC values. Maximum averages of $6.59$, $6.34$, $5.62$, $4.84$, and \mbox{$3.75$\,mm} at a DC of \mbox{$10$\hspace{0.1ex}$\%$} were measured for frequencies of \mbox{$0.5$,~$2$,~$3$,~$4$,~and~$5$\,Hz}, respectively. Photographic composites of the \mbox{peak-to-peak} lateral excursions recorded during these six tests are presented in \mbox{Figs.\,\ref{FIG04}(c)--(h)}. As seen in these images, the resulting stroke envelopes remain approximately symmetric up to the frequency of \mbox{$4$\,Hz}; at \mbox{$5$\,Hz}, the lateral excursion becomes biased toward the right side. As expected, \mbox{Fig.\,\ref{FIG04}(b)} indicates that the mean of the average power consumption increases approximately linearly with the DC value and is mostly independent of the actuation frequency. For a frequency of \mbox{$5$\,Hz}, this relationship can be approximated as \mbox{$\bar{P}_{\ts{s}}(\ts{DC}) = 720\cdot\ts{DC}\,\left[ \ts{mW} \right]$}, with the DC value in per unit. 

\mbox{Fig.\,\ref{FIG05}(a)} summarizes the data obtained from $60$ \mbox{open-loop} \mbox{forward-swimming} tests. In this plot, each data point shows the mean of the measured \mbox{cycle-averaged} forward speeds in \mbox{$10$\,s} of \mbox{steady-state} data and the corresponding ESD. Generally, these results indicate a positive correlation between average \mbox{steady-state} speed and DC for most tested frequencies. Photographic composites of frames, taken at \mbox{$5$-s} intervals, show the tested \mbox{FRISSHBot} prototype swimming at its peak speed for each experimental frequency in~\mbox{Figs.\,\ref{FIG05}(b)--(g)}. The recorded swimming speed with the largest
mean of $13.6\,\ts{mm/s}$ corresponds to a DC value of \mbox{$10$\hspace{0.1ex}$\%$} and a frequency of \mbox{$2$\,Hz}. Notably, the mean speeds corresponding to the same DC of \mbox{$10$\hspace{0.1ex}$\%$}, for frequencies of $3$ and \mbox{$4$\,Hz}, are comparable to the recorded experimental maximum---$11.7$ and $13.1\,\ts{mm/s}$, respectively---while the largest mean values for $0.5$, $1$, and \mbox{$5$\,Hz} are significantly lower, which suggests that the best operating frequencies for achieving high swimming performance are in the range from $2$ to \mbox{$4$\,Hz}. When swimming at its maximum capacity (\mbox{$2$\,Hz}, \mbox{$10$\hspace{0.1ex}$\%$}), the tested \mbox{FRISSHBot} prototype travels about $4.4$ times faster than its prior version, as reported \mbox{in\cite{TrygstadCK2024}}, and achieves CoT, Sw, and St values of \mbox{$9\hspace{0.2ex}304$}, \mbox{$2\hspace{0.2ex}868$}, and $0.93$, respectively. Overall, these values indicate a low efficiency of energy conversion and swimming, as evidenced by the St of $0.93$, which is well outside the reported optimal range of $0.25$ to \mbox{$0.35$\cite{TaylorGK2003}}. The lowest recorded St value of $0.57$ for the tested swimmer occurs when operating at \mbox{$0.5$\,Hz} with a DC of \mbox{$9$\hspace{0.1ex}$\%$}. This St is much closer to the optimal range relative to that achieved at higher speeds; however, the corresponding CoT is \mbox{$20\hspace{0.2ex}127$}. These values indicate that with these excitation parameters, the swimmer is more efficient at transforming its flapping motion into locomotion but significantly more inefficient at transporting its mass, due to the relatively low forward swimming speed of $5.7\,\ts{mm/s}$. Note, however, that this speed is still $1.9$ times larger than that of the original FRISSHBot platform.
\begin{figure*}[t!]
\vspace{1.4ex}    
\begin{center}
\includegraphics[width=0.96\textwidth]{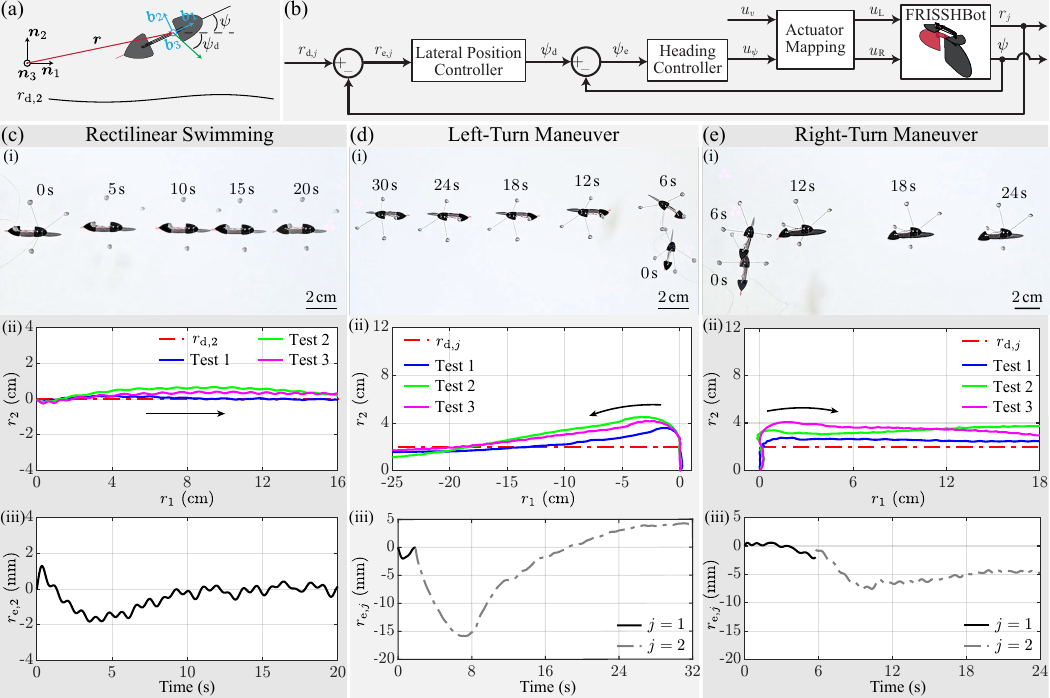}
\end{center}
\vspace{-2ex}
\caption{\textbf{\mbox{Feedback-controlled} swimming experiments of the tested \mbox{FRISSHBot} prototype.}\,\textbf{(a)}\,Definition of the \mbox{laboratory-fixed} inertial and body-fixed frames of reference, \mbox{$\left\{\bs{n}_1,\bs{n}_2,\bs{n}_3\right\}$} and \mbox{$\left\{\bs{b}_1,\bs{b}_2,\bs{b}_3\right\}$}, respectively, and state of the system. Here, $\bs{r}$ is the position of the robot's CoM---which coincides with the origin of the \mbox{body-fixed} frame---relative to the origin of the inertial frame; and, $\psi$ is the swimmer's heading.\,\textbf{(b)}\,Control scheme used during feedback \mbox{trajectory-tracking} swimming tests.\,\textbf{(c)}\,\mbox{Feedback-controlled} \mbox{rectilinear} swimming tests.\,(i)\,Composite of frames taken at \mbox{$5$-s} intervals from video footage of rectilinear \mbox{Test\,1}.\,(ii)\,Measured \mbox{$2$D} trajectory of the swimmer following a \mbox{straight-line} reference path during \mbox{Tests\,1},~2,~and~3. The arrow indicates the direction of motion.\,(iii)\,Lateral control error measured during \mbox{Test\,1}.\,\textbf{(d)}\,\mbox{Feedback-controlled} \mbox{left-turn} tests.\,(i)\,Composite of frames taken at \mbox{$6$-s} intervals from video footage of \mbox{left-turn} \mbox{Test\,1}.\,(ii)\,Measured $2$D trajectories of the swimmer during \mbox{closed-loop} \mbox{left-turn} \mbox{Tests\,1},~2,~and~3. The arrow indicates the direction of motion.\,(iii)\,Lateral control error measured during \mbox{Test\,1}. When the reference path is defined with respect to $r_1$, the control error is computed along that coordinate; consistently, when the reference path is defined with respect to $r_2$, the control error is computed along that coordinate.\,\textbf{(e)}\,\mbox{Feedback-controlled} \mbox{right-turn} tests.~(i)\,Composite of frames taken at \mbox{$6$-s} intervals from video footage of \mbox{right-turn} \mbox{Test\,1}.\,(ii)\,Measured $2$D trajectories of the swimmer during \mbox{closed-loop} \mbox{right-turn} \mbox{Tests\,1},~2,~and~3. The arrow indicates the direction of motion.\,(iii)\,Lateral control error measured during \mbox{Test\,1}. When the reference path is defined with respect to $r_1$, the control error is computed along that coordinate; consistently, when the reference path is defined with respect to $r_2$, the control error is computed along that coordinate. Video footage of these swimming experiments is shown in the accompanying supplementary movie. \label{FIG07}}
\vspace{-2.2ex}
\end{figure*}

\mbox{Figs.\,\ref{FIG06}(a)~and~(b)} summarize the experimental data obtained from \mbox{open-loop} left-turning (\mbox{about\,$\bs{b}_3$}) and right-turning (\mbox{about\,$-\bs{b}_3$}) experiments, respectively. Generally, the turning rate increases with the DC value for both directions---left~and~right. The two best \mbox{left-turning} cases, with average rates of \mbox{$12.0$~and~$10.2$\,\textdegree{/s}}, correspond to the \mbox{$f$--DC} pairs \mbox{$\left\{2\,\ts{Hz},12\text{\hspace{0.1ex}$\%$}\right\}$} and \mbox{$\left\{3\,\ts{Hz},13\text{\hspace{0.1ex}$\%$}\right\}$}, respectively. The two best \mbox{right-turning} cases, with average rates of \mbox{$7.5$~and~$8.9$\,\textdegree{/s}}, correspond to the \mbox{$f$--DC} pairs \mbox{$\left\{4\,\ts{Hz},15\text{\hspace{0.1ex}$\%$}\right\}$} and \mbox{$\left\{5\,\ts{Hz},15\text{\hspace{0.1ex}$\%$}\right\}$}, respectively. Photographic composites of frames, taken at \mbox{$5$-s} intervals, of the \mbox{left-turning} and \mbox{right-turning} cases are presented in \mbox{Figs.\,\ref{FIG06}(c)--(f)}, respectively. Video footage of all the experiments discussed in this section is shown in the accompanying supplementary movie. Overall, the obtained experimental results provide preliminary but compelling evidence that \mbox{single-tail} swimmers of the \mbox{FRISSHBot} type represent a path toward achieving high swimming performance and efficiency at the subgram scale. 

\section{\mbox{Feedback-Controlled} Experiments}
\label{SEC04}
\vspace{-0.5ex}
We performed \mbox{feedback-controlled} \mbox{trajectory-tracking} swimming experiments using the setup depicted in \mbox{Fig.\,\ref{FIG03}(b)}, already described above. The strategy and scheme for \mbox{real-time} control are shown in \mbox{Figs.\,\ref{FIG07}(a)~and~(b)}, which are presented in detail \mbox{in\cite{BlankenshipEK2024}~and~\cite{BenaRM2021}}. As seen, the state of the system is given by the position of the robot's CoM relative to the origin of the inertial frame of reference, \mbox{$\bs{r}$ = $\left[r_1\,\,r_2\right]^T$}, and the heading, $\psi$, defined as the angle between $\bs{b}_1$ and $\bs{n}_1$. As shown in \mbox{Fig.\,\ref{FIG07}(b)}, the control scheme is composed of two controllers:\,(i)\,a \textit{\mbox{lateral-position} controller}~(LPC); and,\,(ii)\,a \textit{heading controller}~(HC). The LPC receives as its input the \mbox{lateral-position} error, \mbox{$r_{\ts{e},j} = r_{\ts{d},j} - r_j$}, where $r_{\ts{d},j}$ and $r_j$ are the desired and measured lateral positions, respectively, with \mbox{$j=1$~or~$2$}, depending on the direction of the desired trajectory---horizontal~or~vertical. The desired heading, $\psi_{\ts{d}}$, is computed by the LPC as 

\vspace{-2ex}
{\small
\begin{align}
\psi_\ts{d}(t) = k_{\ts{p}} r_{\ts{e},j}(t) + k_{\ts{i}}\int_0^t r_{\ts{e},j}(\tau)\,d\tau,
\label{EQN08}
\end{align}
}

\noindent where $k_{\ts{p}}$ and $k_{\ts{i}}$ are proportional and integral controller gains. The HC receives as its input the heading error, \mbox{$\psi_\ts{e} = \psi_\ts{d} - \psi$}, and computes the heading input as \mbox{$u_{\psi} = k_{\ts{p},\psi} \psi_\ts{e}$}, where $k_{\ts{p},\psi}$ is a proportional gain. The actuator mapping receives as its inputs $u_{\psi}$ and the \mbox{open-loop} speed reference, $u_v$, and maps them into DC values for the two PWM signals exciting the robot's bimorph actuator (see the rule presented in\cite{BenaRM2021}).

For \mbox{real-time} implementation, based on the \mbox{open-loop} characterization results discussed in \mbox{Section\,\ref{SEC03}}, we selected the actuation frequency to be \mbox{$3$\,Hz}. The controller gains were empirically chosen to be \mbox{$k_{\ts{p}} = 3\,\ts{rad/m}$}, \mbox{$k_{\ts{i}} = 1\,\ts{rad/} \ts{(m}\cdot \ts{s)}$}, and \mbox{$k_{\ts{p},\psi} = 2\,\ts{/rad}$}; and, the \mbox{open-loop} speed reference was set to \mbox{$u_v = 0.11$} per unit. In this manner, when the heading control error is zero, the two \mbox{SMA-based} subsystems of the driving bimorph actuator are excited using PWM voltages with the same DC value of \mbox{$11$\hspace{0.1ex}$\%$} \mbox{(i.e., $u_{\ts{R}}=u_{\ts{L}}=0.11$)}. When the heading control error is \mbox{nonzero}, the mapping adds $u_{\psi}$ to the DC of the PWM signal exciting one side of the actuator and subtracts the same amount from the DC of the other side, as required to steer the robot in the direction that reduces the trajectory deviation. To avoid any potential actuator damage, we included a saturation law, such that \mbox{$u_{\ts{R}},u_{\ts{L}} \leq 0.22$}, for all time. To assess the swimming performance and maneuverability of the tested robot, we conducted three \mbox{back-to-back} experiments for each of the following modes of operation: rectilinear swimming, right turning, and left turning. The data obtained from these tests are presented in \mbox{Figs.\,\ref{FIG07}(c)--(e)}. Here, \mbox{Subfigures\,(i)} show composites of video frames corresponding to the first test of each type (\mbox{Test\,1}). The frames from the rectilinear swimming test were taken at intervals of \mbox{$5$\,s}; the frames from the turning experiments were taken at intervals of \mbox{$6$\,s}. \mbox{Subfigures\,(ii)~and~(iii)} show the reference and measured trajectories corresponding to \mbox{Tests\,1--3}, and the lateral control error during \mbox{Test\,1}, respectively. In the \mbox{rectilinear-swimming} cases, the tested swimmer achieved lateral control errors with RMS values on the order of \mbox{$2.6$\,mm}, and average swimming speeds in excess of  $9.1\,\ts{mm/s}$. During the \mbox{left-turn} experiments, the tested swimmer achieved an average turning rate of $10.8$\,\textdegree{/s}, an average turning radius on the order of \mbox{$24$\,mm}, and an average RMS value for the tracking error of about \mbox{$10$\,mm}. During the \mbox{right-turn} experiments, the tested swimmer achieved an average turning rate of $13.1$\,\textdegree{/s}, an average turning radius on the order of \mbox{$10$\,mm}, and an average RMS value for the tracking error of about \mbox{$11$\,mm}. These experimental results clearly demonstrate the enhanced capabilities of the new \mbox{FRISSHBot} platform relative to those of the prior version reported \mbox{in\cite{TrygstadCK2024}}.

\section{Conclusion}
\label{SEC05}
\vspace{-0.5ex}
We presented the new \mbox{FRISSHBot}---a \mbox{$59$-mg} \mbox{single-tail} aquatic robot driven by an advanced \mbox{$10$-mg} \mbox{SMA-based} bimorph actuator---which is an evolved \mbox{$2$D-controllable} version of the swimmer first presented \mbox{in\cite{TrygstadCK2024}}. This research was motivated by the goal of creating a bioinspired \mbox{single-muscle} \mbox{mm-scale} swimmer with full controllability and the capability of achieving high swimming performance. The resulting \mbox{physics-informed} design, developed using the notion of resistive drag factor, features an enlarged head and shortened tail relative to those of the original platform. The new swimmer was demonstrated to achieve average forward swimming speeds of up to $13.6\,\ts{mm/s}$~($0.38\,\ts{Bl/s}$) when operating at \mbox{$2$\,Hz}, and high maneuverability---evidenced by the ability to turn at high angular speeds with small radii. During \mbox{feedback-controlled} \mbox{trajectory-tracking} swimming experiments, the tested prototype achieved control errors with RMS values as low as \mbox{$2.6$\,mm}. Along the same lines, the tested swimmer was demonstrated to turn with rates of up to \mbox{$13.1$\,\textdegree{/s}} and turning radii as small as \mbox{$10$\,mm}. To the best of our knowledge, these are the first \mbox{closed-loop} swimming experiments implemented using a \mbox{single-tail} subgram aquatic robot reported to date. Through further design and development, we will continue evolving the current \mbox{FRISSHBot} platform to achieve power and control autonomy, \mbox{high-efficiency} energy conversion, \mbox{high-performance} swimming, and the ability to locomote underwater.   

\bibliographystyle{IEEEtran}
\bibliography{references}
\end{document}